\documentclass[conference]{IEEEtran}
\IEEEoverridecommandlockouts

\usepackage{cite}
\usepackage{amsmath,amssymb,amsfonts}
\usepackage{bm}
\usepackage{algorithmic}
\usepackage{graphicx}
\usepackage{textcomp}
\usepackage{xcolor}
\usepackage{amsmath}
\usepackage{amssymb}

\usepackage{booktabs} 
\usepackage{multirow} 
\usepackage{verbatim}
\usepackage{listings}

\let\P\myP 
\DeclareMathOperator{\NP}{NP}
\def\BibTeX{{\rm B\kern-.05em{\sc i\kern-.025em b}\kern-.08em
    T\kern-.1667em\lower.7ex\hbox{E}\kern-.125emX}}

\usepackage{hyperref}

\begin{document}
\title{P $\approx$ NP, at least in Visual Question Answering
\thanks{This work was supported by NVIDIA's NVAIL program and the BMBF project DeFuseNN (Grant 01IW17002)}}

\author{
\IEEEauthorblockN{
\thanks{\IEEEauthorrefmark{1} Authors contributed equally}
Shailza Jolly\IEEEauthorrefmark{1}\textsuperscript{,1,2},
Sebastian Palacio\IEEEauthorrefmark{1}\textsuperscript{,1,2},
Joachim Folz\textsuperscript{1,2},
Federico Raue\textsuperscript{2},
J{\"o}rn Hees\textsuperscript{2},
Andreas Dengel\textsuperscript{1,2}
}
\IEEEauthorblockA{
\textsuperscript{1}DFKI GmbH\\
\textsuperscript{2}TU Kaiserslautern \\
Kaiserslautern, Germany \\
\url{firstname.lastname@dfki.de}
}
}

\maketitle
\begin{abstract}
In recent years, progress in the Visual Question Answering (VQA) field has largely been driven by public challenges and large datasets.
One of the most widely-used of these is the VQA 2.0 dataset, consisting of polar (``yes/no'') and non-polar questions.
Looking at the question distribution over all answers, we find that the answers ``yes'' and ``no'' account for 38\,\% of the questions, while the remaining 62\,\% are spread over the more than 3000 remaining answers.
While several sources of biases have already been investigated in the field, the effects of such an over-representation of polar vs. non-polar questions remain unclear.

In this paper, we measure the potential confounding factors when polar and non-polar samples are used jointly to train a baseline VQA classifier, and compare it to an upper bound where the over-representation of polar questions is excluded from the training.
Further, we perform cross-over experiments to analyze how well the feature spaces align.

Contrary to expectations, we find no evidence of counterproductive effects in the joint training of unbalanced classes.
In fact, by exploring the intermediate feature space of visual-text embeddings, we find that the feature space of polar questions already encodes sufficient structure to answer many non-polar questions.
Our results indicate that the polar ($\bm{P}$) and the non-polar ($\bm{NP}$) feature spaces are strongly aligned, hence the expression $\bm{P} \bm{\approx} \bm{NP}$.
\end{abstract}

\section{Introduction}
\label{sec:introduction}

The task of Visual Question Answering (VQA) is highly interesting as it requires machine learning (ML) models to concurrently optimize multi-modal objectives related to natural language processing, object detection, and instance segmentation. As initial advancements in the field were quickly made, some research started analyzing the extent to which good results were merely achieved by exploiting low level biases in the datasets themselves. Not long after the release of several VQA datasets, a single study already found consistent significant sources of bias for at least six of them~\cite{b9}.
Said biases are found primarily on the questions, and more precisely, on the way certain questions strongly correlate with the correct answer (e.g., questions that start with ``\textit{what sport...}'' can be answered with ``\textit{tennis}'' more often than not).
Imbalances like these have been widely studied and have even resulted in new metrics~\cite{b9},
regularizers~\cite{b4} or re-balanced partitions for existing datasets~\cite{b7,b4}.

\begin{figure}[t]
\centering
    \includegraphics[width=\linewidth]{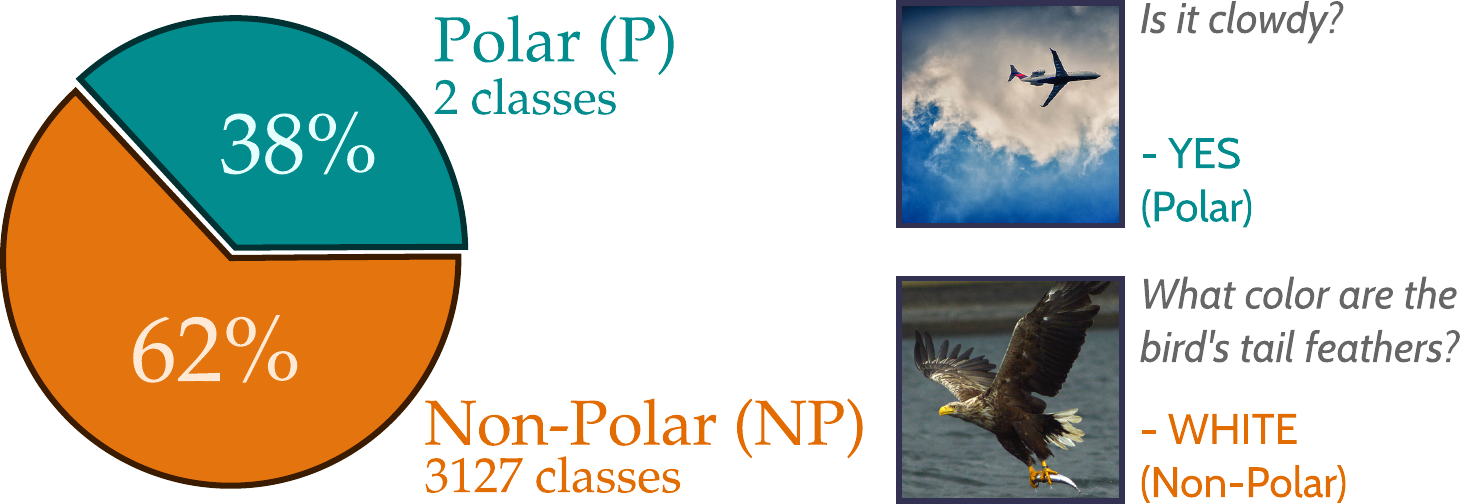}
	\caption{Distribution of polar (P) and non-polar (NP) samples in VQA 2.0.}
	\label{fig:statbias}
\end{figure}

In this paper, we explore a different, potential source of bias in datasets that include both polar ($\P$) and non-polar ($\NP$) questions together.
Polar questions are those with answer \textit{yes} or \textit{no} (commonly referred to as ``yes/no'' questions).
By extension, non-polar questions are the complement of the set of polar questions, i.e., questions with an answer that is something other than \textit{yes} or \textit{no}.
Polar questions have been criticized for their simplicity and often excluded in favor of questions that are richer in complexity~\cite{b10,b14}.
In fact, datasets like COCO-QA, VQA 1.0 or VQA 2.0 that make use of polar questions show an imbalance of the answer distribution where polar questions are significantly over-represented.
For the case of VQA 2.0, polar questions comprise $38\,\%$ of the available corpus.
This means that the answer \textit{yes} appears roughly $19\,\%$ of the time, and so does the answer \textit{no}.
In contrast, each of the remaining 3127 classes (answers) occur, on average, $2\,\%$ of the time (see Figure~\ref{fig:statbias}).

So far, the over-representation of polar samples has been approached in isolation, mainly in two different ways:
First, polar questions are either excluded entirely form the corpus or used exclusively~\cite{b15,b13}.
Second, for datasets with a mix of polar and non-polar questions, evaluation protocol dictates that the accuracy has to be reported separately for polar samples.
Despite this strict separation of the second scenario, VQA models are frequently trained jointly, treating each unique answer independently, regardless of polarity, and under i.i.d.\ conditions.
Moreover, state-of-the-art approaches do not make any specific mention about balancing techniques like class regularization or mini-batch resampling~\cite{b8}.
This simple imbalance can cause severe performance issues, as proposed models could allocate more capacity to answering polar questions just because they appear more often during training.
In fact, we see how performance of polar questions is consistently superior to that of other non-polar sub-categories for the popular VQA challenge\footnote{\url{http://visualqa.org}}.
Even after compensating for the number of classes per group, the expected mean accuracy of any random pair of non-polar classes is a lot lower than the one of polar questions.
The question we ask is if there is any measurable impact (positive or negative) stemming from the over-representation of polar questions in VQA datasets.
Are there confounding factors between polar and non-polar questions when projected into a common feature space?
Are polar questions occupying a non-overlapping region of the feature space w.r.t. their non-polar counterparts?
We investigate these questions and their implications by conducting a series of experiments on a high-performance VQA classifier.
By comparing its behaviour when data distribution changes during training and testing, we conclude, contrary to intuition, that there is a considerable overlap between features from polar questions and non-polar questions.
Moreover, this overlap is favorable to the overall optimization objective such that non-polar questions can be successfully answered based purely on polar features and vice versa.

The contributions of this paper are thereby two-fold:
\begin{itemize}
  \item An evaluation of the potential confounding factors (i.e. bias) that polar questions induce due to over-representation during training.
  \item Empirical evidence indicating that the feature space from polar features can be used to answer non-polar questions and vice versa ($\P \approx \NP$).
\end{itemize}

The rest of the paper is organized as follows.
Section~\ref{sec:relatedwork} goes over previous work related to the study of polar and non-polar questions, as well as biases in VQA datasets.
Section~\ref{sec:methods} outlines the experimental setup and training regimes.
Section~\ref{sec:results} describes the experimental results and Section~\ref{sec:discussion} brings results into perspective, discussing the implications in the context of the joint feature space of polar and non-polar questions.

\section{Related Work}
\label{sec:relatedwork}

Current developments in the field of deep learning have demanded large amounts of data for training and evaluation.
Models rely on the usefulness of these carefully curated corpuses, often taking for granted how representative a training set is.
Datasets being permeable to biases (spurious, undesired patterns) can drastically undermine the novelty and usefulness of certain ML models, as well as their claimed performance.

Problems dealing with natural text often struggle with issues of this nature.
For example, a strong gender bias has been reported for imSitu, a visual semantic role labeling dataset where activities like ``\textit{cooking}'' were strongly biased towards women~\cite{b6}.
Not only was the bias present in the dataset but trained models were amplifying the bias during testing as well.
The widely used MS-COCO dataset~\cite{b16} presents biases for image captioning because objects and background often co-occur e.g., giraffes appear next to a tree with grass in the background~\cite{b13}.
Being largely based on MS-COCO, the VQA 1.0 dataset~\cite{b3} was therefore affected by this bias too, with the now infamous example of ``\textit{tennis}'' being the correct answer to 41\% of questions starting with ``\textit{what sport...}''.
Not long after, a study focusing on the limitations of current VQA datasets found that six of the most commonly used corpuses contained some sort of bias related to the textual domain~\cite{b9}.
The ubiquity of said imbalances results in an inability to measure the extent by which VQA models are indeed capable of extracting meaningful, and visually grounded semantics.

From this point on, several advances in VQA were directly addressing biases in the existing datasets.
Alternative training and testing splits for VQA 1.0 and 2.0 were introduced in order to measure the extend by which VQA models can cope with unseen composition of concepts e.g., after learning about ``green plate'' and ``white shirt'' the network is then evaluated on ``white plates'' or ``green shirts''~\cite{b7,b4}.
Simultaneously, the emergence of several regularization techniques helped models compensate or at least mitigate the effects of biases picked up by language models~\cite{b5,b4}.

On the behavior of polar and non-polar questions, research has focused, among other things, on the importance of having polar questions balanced, i.e., having the same number of questions with ``\textit{yes}'' and ``\textit{no}'' as an answer~\cite{b13}.
Moreover, a polar-only dataset was proposed to measure the extent by which logic reasoning can be solved by a VQA system~\cite{b12}.
So far, these experiments only analyze effects of imbalances within the polar space in isolation, disregarding the interactions that polar and non-polar questions may have in a joint feature space.
A recapitulation of good practices for training models using VQA 2.0 recommends balancing each mini-batch w.r.t. opposite questions~\cite{b2}.
This policy ensures that there is always a balance between samples with ``\textit{yes}'' and ``\textit{no}'' answers, but does not consider the balance between polar and non-polar questions.
An explicit separation of polar and non-polar questions was proposed for the GVQA model, which processes polar questions separately from non-polar ones, using the former as a verification mechanism for the non polar concepts contained within~\cite{b4}.

Since polar questions can be obtained easier than non-polar ones, datasets relying on human annotators end up with a distribution of question types that is heavily skewed towards polar questions.
Some synthetically generated datasets like CLEVR compensate for such over-representation, having a more uniform distribution along different answer types~\cite{b11}.
On the other side of the spectrum, synthetic and natural datasets have outright dismissed the use of polar questions to alleviate such over-representation issues~\cite{b14}, and because of this restriction, more complex questions can be attained~\cite{b10}.

In this paper, we are concerned with the influence that over-represented polar questions exert on the non-polar counterparts when trained jointly, as it is the case for a majority of modern VQA models.
Experiments that eliminate confounding factors between the two categories are conducted in order to assess the upper bound of a VQA model when using only one or the other kind of question.
Furthermore, we find a strong alignment between the feature space of polar questions and that of non-polar questions, indicating that polar questions can be used to answer non-polar answers and vice versa.

\section{Methods}
\label{sec:methods}

In this section, we present the main model and variations thereof used for all the experiments, motivate the use of the dataset, and describe the metrics to be compared afterwards.

\subsection{Dataset}
\label{sec:dataset}
We use the VQA 2.0 dataset for all experiments.
With $443\,757$ training samples and $214\,354$ for validation, it is currently the largest available non-synthetic corpus for VQA containing both polar and non-polar questions.
Additional properties that make this dataset suitable for our analysis include a uniform distribution between questions with ``\textit{yes}'' and ``\textit{no}'' as ground-truth, as well as an adjusted distribution of non-polar answers w.r.t.\ VQA 1.0.
We split the VQA 2.0 corpus into two disjoint sets corresponding to the polar and non-polar questions.
Notwithstanding, samples also remain in the training and validation set as originally assigned.

\subsection{VQA Reference Model}
\label{sec:vqamodel}
\begin{figure}[t]
	\includegraphics[width=\linewidth]{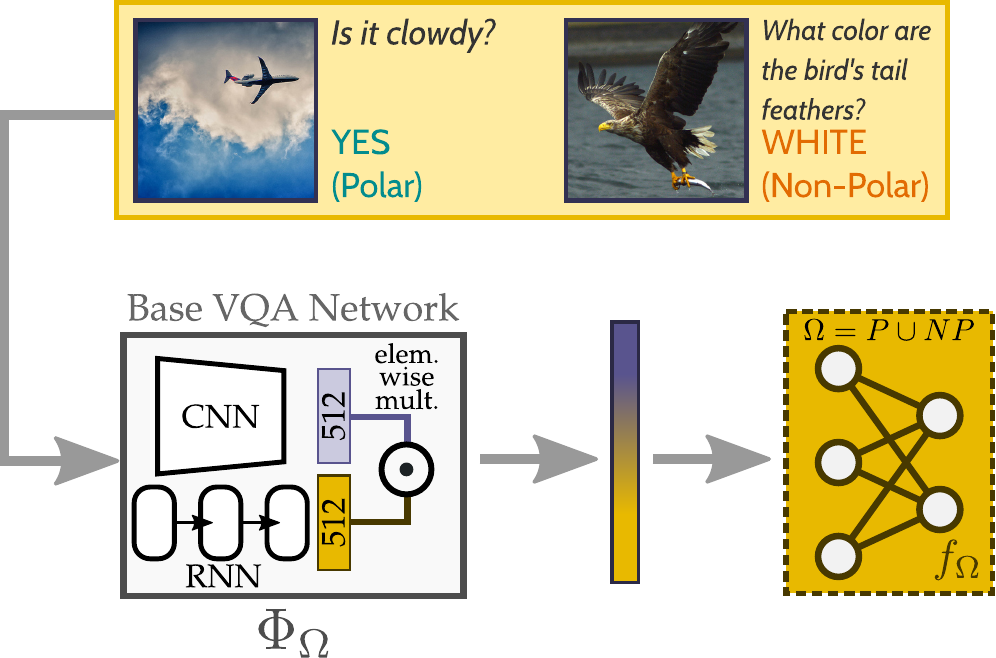}
	\caption{Overview of the VQA model used throughout this paper. This is an re-implementation from the winning entry of the 2017 VQA challenge. It is composed by two main modules: a textual-visual joint embedding (base VQA model denoted as $\Phi_{\Omega}$) and a shallow 2-layer classifier (denoted as $f_{\Omega}$).}
	\label{fig:designarch}
\end{figure}

For our experiments, we use a high-performance VQA system with region-based attention~\cite{b1}.
Figure~\ref{fig:designarch} shows an overview of the model which, at its core, consists of three modules: an image embedding, a text embedding and a joint classification module.
To model the joint visual-text space, both image and text embeddings are first projected to a 512-dimensional space, and then fused together through an element-wise product.
Subsequently, the joint embedding passes through a sequence of fully connected layers before reaching the output layer, where the output is normalized by a softmax operation.
In this paper, we refer to the first part of the network (up until the point-wise multiplication of the 512-dimensional projection of the visual-text space) as the \textbf{base VQA network}, denoted as $\Phi$.
The remaining part of the network consisting of two fully connected layers and the output layer is referred to as \textbf{the classifier} and denoted as $f$.
This way, a prediction by the network can be written as $\hat{y} = f(\Phi(x))$.
When only using polar questions for training the model, we use $\Phi_{\P}$ and $f_{\P}$ to denote the corresponding modules.
Similarly, $\Phi_{\NP}$ and $f_{\NP}$ refer to modules that have been trained using non-polar samples only.
For completeness, $\Phi_{\Omega}$ and $f_{\Omega}$ denote modules trained on both polar and non-polar questions. 

\subsection{Experiments}
\begin{figure}[t]
	\includegraphics[width=\linewidth]{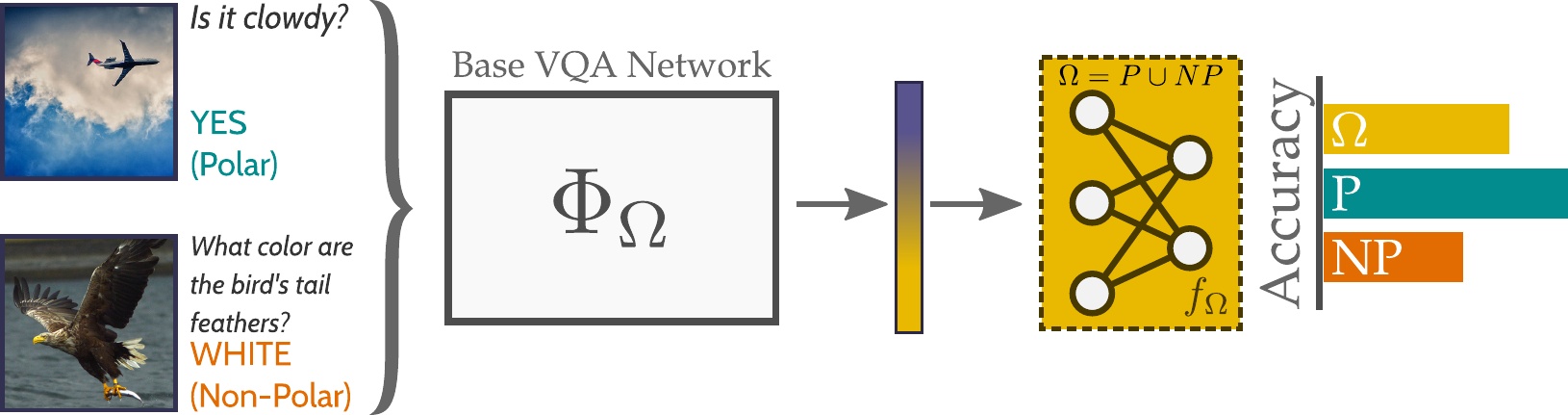}
	\caption{Outline of the baseline experiment: The full VQA model is trained on the full VQA 2.0 training set. Accuracy is reported on the full validation set $\Omega$, on the polar questions in the validation set $\P$ and on the non-polar counterparts $\NP$.}
	\label{fig:baseline}
\end{figure}

There are three main experiments in this work, and they can be summarized as follows:

\textbf{Baseline:} We train the entire model on VQA 2.0 without any additional considerations regarding polar and non-polar questions.
As in the original work~\cite{b1}, we make use of samples from VQA 2.0 containing answers that appear at least eight times in the entire dataset. 
This yields an answer space of 3129 dimensions, two of which correspond to the classes ``\textit{yes}'' and ``\textit{no}'' (i.e., the polar space).
No additional pre-training from Visual Genome is used to avoid potential effects of biases coming from another dataset.
The number of regions used for the image embedding is fixed at 36.
Our final implementation uses ReLUs instead of the originally proposed GatedTanh, since it saves computation and produces similar results. 
The model is trained using Adamax~\cite{b17} with an initial learning rate of $2\times 10^{-3}$ on the full training set, and the standard VQA accuracy is reported for the validation set~\cite{b3}.
We report the VQA accuracy of the baseline w.r.t.\ three splits of the validation set: 1) using only the polar questions in the validation set, 2) using only the non-polar questions and 3) passing the entire validation set which contains both polar and non-polar questions (see Figure~\ref{fig:baseline}).
The baseline is used as a reference to quantify the impact of polar and non-polar questions when used together during training (this experiment) or separately (following experiments).

\begin{figure}
	\includegraphics[width=\linewidth]{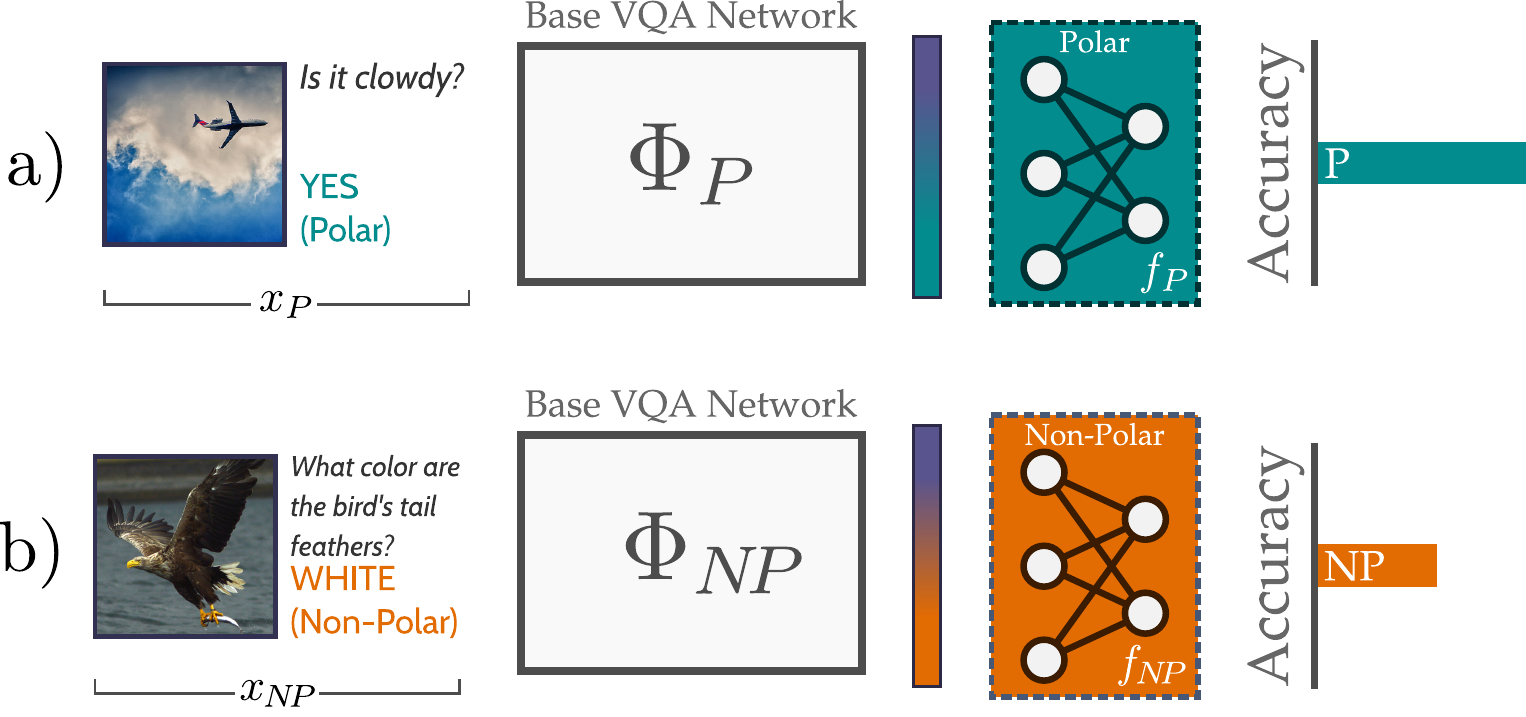}
	\caption{Unbiased Upper Bound experiment: two copies of the same architecture for the baseline is used. One copy is trained only on polar questions and the second copy is trained only on non-polar questions. Accuracy for both is reported independently.}
	\label{fig:bestline}
\end{figure}

\textbf{Unbiased Upper Bound:} To get an empirical upper bound of the model, where the issue of over-representation does not play a role, we train two separate versions of the same model from scratch.
First, we train a model only using the polar questions.
Then we train a second model only using the non-polar questions.
We refer to these two models as $f_{\P} \circ \Phi_{\P}$ and $f_{\NP} \circ \Phi_{\NP}$ respectively.
For these two models, their corresponding VQA accuracy is reported (see Figure~\ref{fig:bestline}): $f_{\P} \circ \Phi_{\P}$ is evaluated on the polar questions of the validation set, and $f_{\NP} \circ \Phi_{\NP}$ on the non-polar ones.
The purpose of this setup is to compare the capacity of the same VQA model used for the baseline, when dealing only with one kind of question.
By training on polar or non-polar questions only, the network can use $100\,\%$ of its capacity to extract the necessary semantics without the burden of modeling features to distinguish between polar and non-polar questions.
In other words, any bias caused by the imbalance of polar and non-polar questions is excluded for these two models.
The accuracy for both variants is thereby expected to be higher than the corresponding value for the baseline experiment.
If small or no deviations arise w.r.t. the baseline, then we can conclude that the confounding factors between polar and non-polar questions are not affecting the baseline VQA model negatively.

\begin{figure*}
	\includegraphics[width=\textwidth]{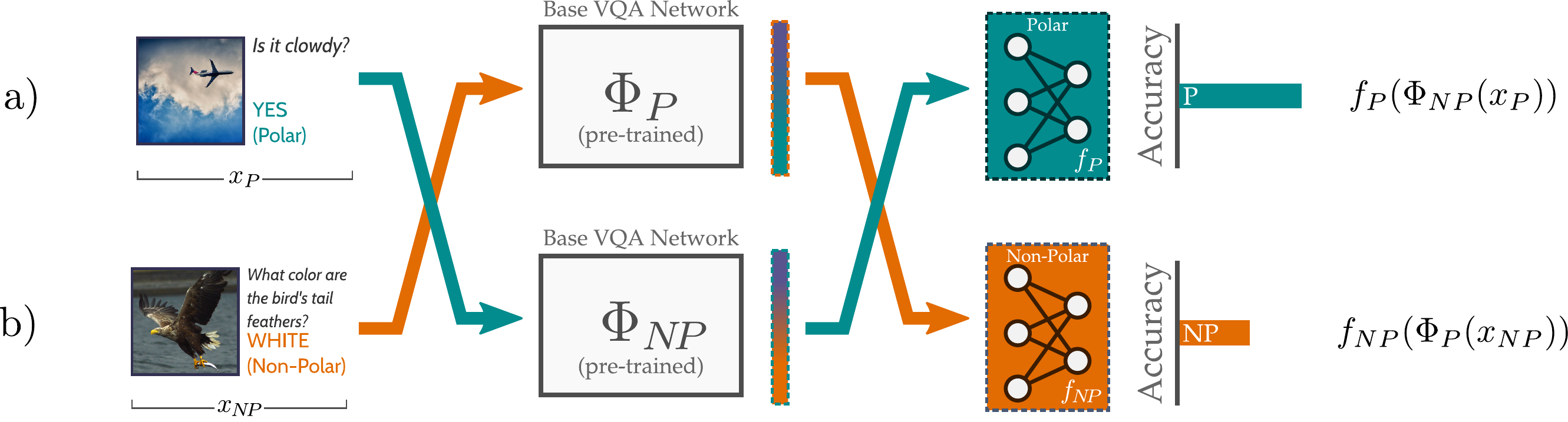}
	\caption{Cross-Polarity evaluation:
    	a) We project polar inputs $x_{\P}$ using a pre-trained base network $\Phi_{\NP}$, which has only seen non-polar samples during its training, into non-polar feature space $\Phi_{\NP}(x_{\P})$ and train a shallow polar classifier $f_{\P}(\Phi_{\NP}(x_{\P}))$ on this representation.
    	b) Vice-versa.
    	Intuitively, the experiment measures the extent by which non-polar questions can be answered based on features extracted from a polar space and vice versa.}
	\label{fig:crossover}
\end{figure*}

\textbf{Cross-Polarity Evaluation:} Independent of the potential confounding factors between polar and non-polar samples, there is still the question on how the distribution of polar features overlap with that of the non-polar ones.
As the feature projections from $\Phi_{\P}$ and $\Phi_{\NP}$ share the same dimensional space, an experiment using transfer learning can be performed.
The outline of this experiment is shown in Figure~\ref{fig:crossover}.
First, we use $\Phi_{\P}$ and $\Phi_{\NP}$ from the previous experiment as fixed pre-trained feature extractors.
Then, we train a new polar classifier $f_{\P}$ using features from the fixed pre-trained non-polar module $\Phi_{\NP}$.
Similarly, we train a new non-polar classifier $f_{\NP}$ with features coming from $\Phi_{\P}$.
In order to be able to compare the results, we use the same architecture for $f_{\P}$ and $f_{\NP}$ as in the previous unbiased upper bound experiments.
Intuitively, this cross-over experiment helps us measuring how descriptive the feature space of polar questions is, so that non-polar questions can also be answered, and vice versa.
At a semantic level, this is theoretically possible since non-polar questions can be asked using a polar structure.
E.g., the question ``\textit{What color are the bird's tail feathers?}'' with answer ``\textit{white}'' can be transformed into the polar question ``\textit{Are the bird's tail feathers white?}''.
Intuitively, it is expected that the space of non-polar concepts (i.e., the joint visual-text projection trained on the non-polar questions $\Phi_{\NP}$ covering more than 3000 answer classes) represents a rich enough structure that can be condensed and reused to answer polar questions (just 2 answer classes).
However, it is not expected that the set of polar-questions yields a rich enough embedding space covering the wide spectrum of non-polar concepts found in VQA 2.0.

\section{Results}
\label{sec:results}

\begin{table}
\centering
\caption{Summary of experimental results. The second column indicates the data used to train each of the VQA modules $\Phi$ and $f$. The column ``\textit{Input}'' indicates the data used during evaluation of the ensemble $f(\Phi(x))$, and the column ``\textit{Accuracy}'' reports the corresponding single-model VQA accuracy~\cite{b3} from the validation set.}
\label{tab:results}
\begin{tabular}{@{}lllll@{}}
\toprule
Task                 & \multicolumn{2}{c}{Model}                     & Input  & Accuracy\\ 
\cmidrule(lr){2-3}
                     & $\Phi$             & $f$                       &  $x$     &      \\
\midrule
Random Choice        & --                 & --                        & $\P$      & 0.5   \\
                     & --                 & --                        & $\NP$     & $0.0003$\\ 
\midrule
                     & {\multirow{3}{*}{$\Omega$}} & {\multirow{3}{*}{$\Omega$}} & $\Omega$ & 0.624 \\
Baseline             & \multicolumn{2}{c}{}                          & $\P$      & 0.804 \\
                     & \multicolumn{2}{c}{}                          & $\NP$     & 0.514 \\ 
\midrule
Upper bound          & $\P$                & $\P$                       & $\P$      & 0.796 \\
                     & $\NP$               & $\NP$                      & $\NP$     & 0.516 \\ 
\midrule
Cross-Polarity       & $\NP$               & $\P$                       & $\P$      & 0.758 \\
                     & $\P$                & $\NP$                      & $\NP$     & 0.287 \\
\bottomrule
\end{tabular}
\end{table}

In this section, we report the results from experiments described in Section~\ref{sec:methods}.

Table~\ref{tab:results} gives an aggregated overview of the main experimental results from all three proposed experiments.
The columns show, from left to right, the name of the experiment, the subset ($\P$: Polar; $\NP$: Non-polar; $\Omega$: All) of the set of samples used for training the corresponding module of the whole VQA ensemble ($\Phi$ or $f$), the subset of the validation set used for evaluation, and the resulting VQA accuracy.
For the cross-polarity experiments, $\Phi$ is assumed to be pre-trained and fixed, and only the corresponding $f$ has been trained from scratch.

We observe that the baseline experiment reaches an accuracy that is within one percentage point of the one reported by the original authors~\cite{b1}. 
The decisions to not involve another dataset for pre-training (e.g., the Visual Genome as in the original work) naturally affects the overall accuracy, but allows the results presented here to reflect more closely the behavior of the polar and non-polar disparity, while ruling out other potential sources of bias.

For the characterization of the upper bound, we see that the reached accuracy falls almost exactly within the range of the baseline.
Upper bound results show an increase of $0.2\,pp$ for a system trained on non-polar questions while training with polar questions decreases accuracy by $0.8\,pp$.
These negligible fluctuations between the upper bound and the baseline strongly suggest that no confounding factors exist between polar and non-polar samples when trained jointly.
In fact, we see that polar questions rarely get confused with any non-polar alternative (Table~\ref{tab:confmatrix}).
Similarly, non-polar questions are not frequently mistaken for any of the two polar answers.
For further in-depth analysis, please refer to the discussion in Section~\ref{sec:discussion}.

\begin{table}
\centering
\caption{Confusion matrix of predictions for the baseline model, grouped by polarity (all numbers in percent). Polar predictions are rarely confused by any of the non-polar alternatives and vice versa.}
\label{tab:confmatrix}
\begin{tabular}{@{}cc|cc@{}}
\multicolumn{1}{c}{} &\multicolumn{1}{c}{} &\multicolumn{2}{c}{Predicted Answer} \\ 
\multicolumn{1}{c}{} & 
\multicolumn{1}{c|}{} & 
\multicolumn{1}{c}{$\P$} & 
\multicolumn{1}{c}{$\NP$} \\ 
\cline{2-4}
\multirow{2}{*}{True Answer}
& $\P$   & 37.59             & \phantom{0}0.11   \\[1.5ex]
& $\NP$  & \phantom{0}0.77   & 61.53             \\ 
\cline{2-4}
\end{tabular}
\end{table}

The cross-polarity experiments exhibit a slightly different behaviour.
First, training a polar classifier $f_{\P}$ based on non-polar features from $\Phi_{\NP}$ yields results almost as high as when $\Phi_{\P}$ is used.
Only $3.8\,pp$ of accuracy separate the cross polar model $f_{\P} \circ \Phi_{\NP}$ and the polar upper bound $f_{\P} \circ \Phi_{\P}$.
Taking into account that both results are above $75\,\%$ accuracy, their difference can be considered low, especially when compared against the probability of randomly guessing which lies at $50\,\%$.
Second, the inverse cross-polar experiment, where non-polar questions are classified using a polar-only feature space $f_{\NP} \circ \Phi_{\P}$ shows an accuracy of $28.7\,\%$.
This presents a loss of $-22.9\,pp$ w.r.t.\ the upper bound $f_{\NP} \circ \Phi_{\NP}$.
Note that this result still lies notably above random chance (which corresponds to $0.03\,\%$), and will be discussed further in the next section.

\section{Discussion}
\label{sec:discussion}

In this section, we analyze the results from all conducted experiments and their implications, in the context of the joint space shared by polar and non-polar features.

The first observation comes from comparing the baseline experiments against their corresponding empirical upper bound for both polar and non-polar features.
Having almost indistinguishable results when the model is being trained with non-polar and over-represented polar samples together, indicates that the model is capable of coping with both question types simultaneously, without compromising performance.
In essence, this simple comparison shows no measurable confounding factors (i.e., source of bias) by populating the visual-text feature space with polar and non-polar questions at the same time.
The need for balancing strategies, commonly used for long tail datasets like mini-batch resampling or weighted labels~\cite{b8} are thereby unlikely to improve the performance of typical VQA model architectures like the one used in this work.
Regarding the distribution of polar and non-polar questions in the joint feature space $\Phi_\Omega$, we are left with two possible scenarios: either each distribution occupies a different (disjoint) sub-region of the feature space or they overlap and hence, they (at least partially) model the same semantic concepts.
An analysis of the remaining experiments will help identifying which of these two conjectures can be verified.

The second observation comes from the first cross-polarity experiment.
Here, by training a classifier on polar questions that are first projected to a feature manifold of non-polar concepts, we measure how possible it is to answer polar questions by using the feature space of non-polar concepts.
This scenario is intuitively simple, because of the vast complexity of topics covered by non-polar questions (3127 answer classes in our experiments).
Also, by reducing the number of classes from 3127 to 2, the chance level of the classification problem increases from $\tfrac{1}{3127}$ to $\tfrac{1}{2}$, which makes the problem much easier.
Hence, the observed result is not really surprising: with only a minor drop in performance, we can say that most polar questions can be answered even by using a non-polar feature space representation.
Looking for the alignment between a non-polar concept (e.g., ``\textit{green}'' or ``\textit{bicycle}'') and the occurrence of that concept in the question embedding, makes polar questions straightforward to answer, even by a simple classifier.

The third, and perhaps most interesting observation comes from the other direction of the cross-polarity experiments.
In this experiment we trained a classifier on non-polar questions while relying on a polar feature space.
Unlike before, we are now going up from 2 to 3127 classes, intuitively making the problem much more challenging.
Also it is unclear to what extent we can expect the polar feature space to be able to express the intricacies of non-polar questions.
As mentioned in Section~\ref{sec:relatedwork}, several VQA datasets have decided to entirely leave out polar samples to favor the more complex questions arising from only allowing non-polar queries.
According to these premises, we expect the non-polar classifier based on features from a polar embedding space to perform poorly.
Recall that the reported accuracy for this setup is $28.7\,\%$ which is indeed lower than the upper bound of $51.6\,\%$.
However, we note that this value is still significantly higher than random chance of $\tfrac{1}{3127}$.
This behavior suggests that there is a notable subset of non-polar questions that can be answered with high accuracy based only on the feature space of polar questions, immediately leading to the question:
How to find the subset of non-polar questions that can be answered through the polar feature space?
The cross-polarity experiments already show that non-polar questions with numeric answers are well conveyed by the polar feature space (accuracy of $f_{\NP} \circ \Phi_{\P}$ for numeric labels such as ``0'', ``1'' and ``2'' is high).
We theorize that a general alignment between polar questions \emph{about} non-polar concepts and the corresponding non-polar questions exists.

\begin{figure}
	\includegraphics[width=\linewidth]{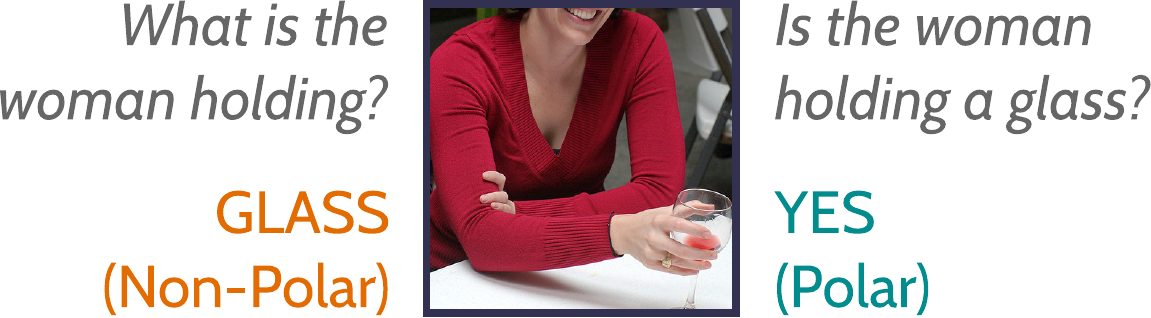}
	\caption{Polar questions can be used to answer non-polar questions with high accuracy as long as the polar questions relate to an existing non-polar concept. Given the image in the center, a polar question (right) and a non-polar question (left) can be asked about a common non-polar concept, namely ``\textit{glass}''.}
	\label{fig:npconcept}
\end{figure}

\subsubsection{Polar Questions About Non-Polar Concepts}
Take the example in Figure~\ref{fig:npconcept}: The non-polar question ``\textit{What is the woman holding?}'' and the the polar question ``\textit{Is the woman holding a glass?}'' relate to the same semantic concept, namely ``\textit{glass}''.
By counting the number of polar questions that talk about each of the non-polar concepts, we can focus on the non-polar concepts which appear in most polar questions.
We call this category of non-polar answers ``well covered''.
We use the notation $\mathcal{X}_{\NP'}$ to refer to non-polar samples (i.e., questions and answers) with answers that are well covered by polar questions.
The complement of $\mathcal{X}_{\NP'}$ (i.e., non-polar samples whose answers are not well covered by polar questions) is denoted by $\overline{\mathcal{X}_{\NP'}}$.
We can then test the $f_{\NP} \circ \Phi_{\P}$ model used in the last experiment of the cross-polarity evaluation w.r.t. $\mathcal{X}_{\NP'}$ and $\overline{\mathcal{X}_{\NP'}}$ separately.

\begin{figure}
	\includegraphics[width=\linewidth]{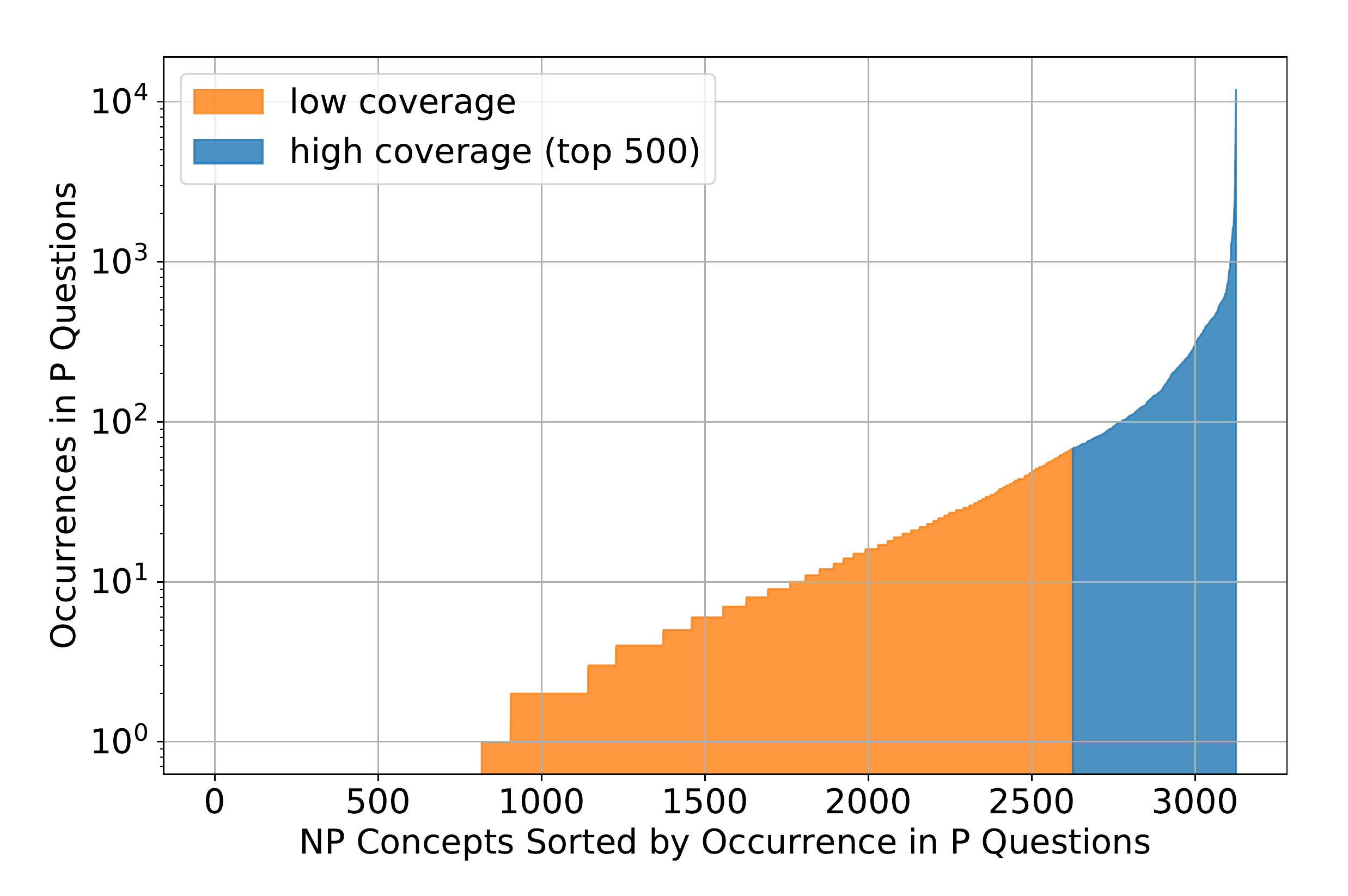}
	\caption{Histogram of non-polar concepts that appear in polar questions within the VQA 2.0 dataset. The 500 non-polar concepts with most matching polar questions define the set $\mathcal{X}_{\NP'}$ (plotted in blue). In other words, the set $\mathcal{X}_{\NP'}$ contains all non-polar samples whose answers have the highest textual occurrences in polar questions.}
	\label{fig:npcoverage}
\end{figure}

To populate the subset $\mathcal{X}_{\NP'}$, we start by selecting polar questions in which any of the 3127 non-polar answers occur textually using a simple regular expression. 
We then count the number of polar question occurrences for each non-polar answer (with replacement) and sort them in ascending order.
The resulting histogram is shown in Figure~\ref{fig:npcoverage}.
The x-axis represents the 3127 non-polar answers and the y-axis represents the number of polar questions matching the non-polar answer.
We see that 73.87\% of non-polar concepts are matched by at least 1 polar question.
To guarantee that each non-polar concept in the subset is covered by a large number of polar questions, we select the top 500 non-polar answers that occur the most often within polar questions (i.e., the 500 best covered non-polar answers) to assign non-polar samples to $\mathcal{X}_{\NP'}$ (shown in blue in Figure~\ref{fig:npcoverage}).
Once $\mathcal{X}_{\NP'}$ (500 classes), and thereby $\overline{\mathcal{X}_{\NP'}}$ (2627 classes) have been defined, we use them to evaluate $f_{\NP} \circ \Phi_{\P}$ separately.
Results of this experiment are shown in Table~\ref{tab:lastcross}.

\begin{table}
\centering
\caption{Accuracy on a cross-polarity experiment where the base VQA feature extractor is either pre-trained on $\mathcal{X}_{\P'}$ (the set of polar questions matching the 500 non-polar concepts) or $\overline{\mathcal{X}_{\P'}}$ (the complement of $\mathcal{X}_{\P'}$).}
\label{tab:lastcross}
\begin{tabular}{@{}lllll@{}}
\toprule
Task                 & \multicolumn{2}{c}{Model}                     & Input  & Accuracy\\ 
\cmidrule(lr){2-3}
                     & $\Phi$             & $f$                       &  $x$             &      \\
\midrule
Cross-Polarity       & $\P$                  & $\NP$                    & $\NP'$            & 0.40  \\
                     & $\P$                  & $\NP$                    & $\overline{\NP'}$ & 0.14  \\ 
\bottomrule
\end{tabular}
\end{table}

When evaluating the polar feature space w.r.t.\ well-covered non-polar questions, the VQA model exhibits an ample improvement (from $28.7\,\%$ to $40\,\%$) compared to the initial cross-polar experiment from Section~\ref{sec:results}.
These results are in turn, closer to the upper bound of non-polar questions that can be answered with the VQA model used throughout all experiments.
In contrast, the polar feature space of poorly covered non-polar concepts presents a steep decrease in accuracy, from $28.7\,\%$ to $14\,\%$.
These results give a strong indication that non-polar questions can be answered by using a feature space based on polar samples.
The caveat is, quite naturally, that the set of polar questions used for training, has to convey enough semantics about the corresponding non-polar questions.

In light of these results, we construct a logical argument, where a sufficient condition for modeling a non-polar questions with a polar feature space depends on having polar questions that deal with the corresponding non-polar concepts.
Therefore, we arrive at the conclusion that the polar feature space \textit{can} carry an equivalent semantic value as the non-polar feature space, hence $\P \approx \NP$.

\section{Conclusion and Future Work}
\label{sec:conclusions}

In this paper, we have presented an in-depth evaluation of the influence that polar and non-polar questions exert on each other when used jointly for training a VQA system.
The over-representation of polar samples w.r.t.\ the non-polar counterparts poses two main questions which we developed throughout this work: (1) Are there any source of bias stemming from the polar and non-polar imbalance and (2) what relationship exists between polar and non-polar samples when projected into the joint feature space that they are usually represented in?

On the first question, we found no confounding factors from the imbalance of polar and non-polar questions, and thereby no detrimental source of bias which may require special attention by doing class weighting or mini-batch re-sampling.

On the second question, we establish a clear correlation between the distribution of polar and non-polar feature embeddings.
We show that polar features can be used to answer non-polar questions, provided that the polar questions used for training refer to the semantic concepts being considered in the non-polar questions.
Based on these findings we conclude that the space of polar features ($\P$) provides a rich semantic structure, similar to that of the non-polar counterparts ($\NP$).
We use the expression $\P\approx \NP$ to refer to this alignment.

\subsubsection{Future Work} the results of this work indicate that the problem of visual question answering for non-polar concepts can be solved using polar questions, as long as polar questions cover the relevant non-polar topics.
We are interested in measuring the empirical extent by which this phenomenon holds.
The usefulness of a feature space based on polar samples for answering non-polar questions is not only surprising, but also potentially ground-breaking because it can change the way future VQA datasets are compiled.
Given the reduced cost of collecting polar questions (compared to non-polar questions), crowd sourcing efforts to amass a critical amount of polar questions for VQA 2.0 could benefit from our findings.
This will allow us to bridge the gap between non-polar concepts that are not well covered by polar questions and complement today's training data.
We also want to explore automatic means to turn non-polar questions into polar ones using natural-language-processing tools.

Furthermore, we are interested in measuring the extent to which a growing number of non-polar concepts can be modeled in the joint visual-text space by only using polar input samples.
Therefore, an arbitrary number of concepts could be explicitly imposed in the joint feature space while keeping a fixed 2-dimensional classification objective. 
This training regime, resembles the conditions of generative adversarial networks (GANs), and could open the possibility to learn new classes over time, which has potential applications in the field of continuous learning.

\end{document}